\documentclass[11pt,a4paper]{article}
\usepackage[utf8]{inputenc}
\usepackage[hyperref]{acl2020}
\usepackage{times}
\usepackage{latexsym}
\usepackage{amsmath}
\usepackage{booktabs}
\usepackage{graphicx}
\usepackage{multirow}
\usepackage{microtype}
\usepackage{url}

\aclfinalcopy 

\newcommand\bQ{\mathbf{Q}}
\newcommand\bK{\mathbf{K}}
\newcommand\bV{\mathbf{V}}
\newcommand\bP{\mathbf{P}}
\newcommand\bS{\mathbf{S}}
\newcommand\bR{\mathbf{R}}
\newcommand\bT{\mathbf{T}}
\newcommand\bM{\mathbf{M}}
\newcommand\bD{\mathbf{D}}
\newcommand\bW{\mathbf{W}}

\newcommand\be{\mathbf{e}}
\newcommand\bm{\mathbf{m}}

\title{ScriptWriter: Narrative-Guided Script Generation}

\author{Yutao Zhu$^{1}$, Ruihua Song$^{2,}$\thanks{$^*$Corresponding authors.} , Zhicheng Dou$^{3,}$\footnotemark[1] , Jian-Yun Nie$^{1}$, Jin Zhou$^{4}$ \\
  $^{1}$Université de Montréal, Montréal, Québec, Canada \\
  $^{2}$Microsoft XiaoIce, Beijing, China \\
  $^{3}$Gaoling School of Artificial Intelligence, Renmin University of China, Beijing, China \\
  $^{4}$Beijing Film Academy, Beijing, China \\
  \texttt{yutao.zhu@umontreal.ca, rsong@microsoft.com} \\
  \texttt{dou@ruc.edu.cn, nie@iro.umontreal.ca, whitezj@vip.sina.com} \\
  \\
}

\date{}

\begin{document}
\maketitle
\begin{abstract}
It is appealing to have a system that generates a story or scripts automatically from a storyline, even though this is still out of our reach. In dialogue systems, it would also be useful to drive dialogues by a dialogue plan. 
In this paper, we address a key problem involved in these applications -  guiding a dialogue by a narrative. The proposed model ScriptWriter selects the best response among the candidates that fit the context as well as the given narrative. It keeps track of what in the narrative has been said and what is to be said. 
A narrative plays a different role than the context (i.e., previous utterances), which is generally used in current dialogue systems.
Due to the unavailability of data for this new application, we construct a new large-scale data collection \textit{GraphMovie} from a movie website where end-users can upload their narratives freely when watching a movie. Experimental results on the dataset show that our proposed approach based on narratives significantly outperforms the baselines that simply use the narrative as a kind of context.
\end{abstract}
\section{Introduction}
Narrative is generally understood as a way to tell a story. WordNet defines it as ``a message that tells the particulars of an act or occurrence or course of events; presented in writing or drama or cinema or as a radio or television program"\footnote{\url{http://wordnetweb.princeton.edu/perl/webwn?s=narrative}}.
Narrative plays an important role in many natural language processing (NLP) tasks. For example, in storytelling, the storyline is a type of narrative, which helps generate coherent and consistent stories~\cite{fan-etal-2018-hierarchical, DBLP:conf/acl/FanLD19}. In dialogue generation, narrative can be used to define a global plan for the whole conversation session, so as to avoid generating inconsistent and scattered responses  ~\cite{DBLP:XingWWHZ18,DBLP:conf/acl/TianYMSFZ17,DBLP:conf/aaai/GhazvininejadBC18}.

\begin{figure}[!t]
    \centering
    \includegraphics[width=\linewidth]{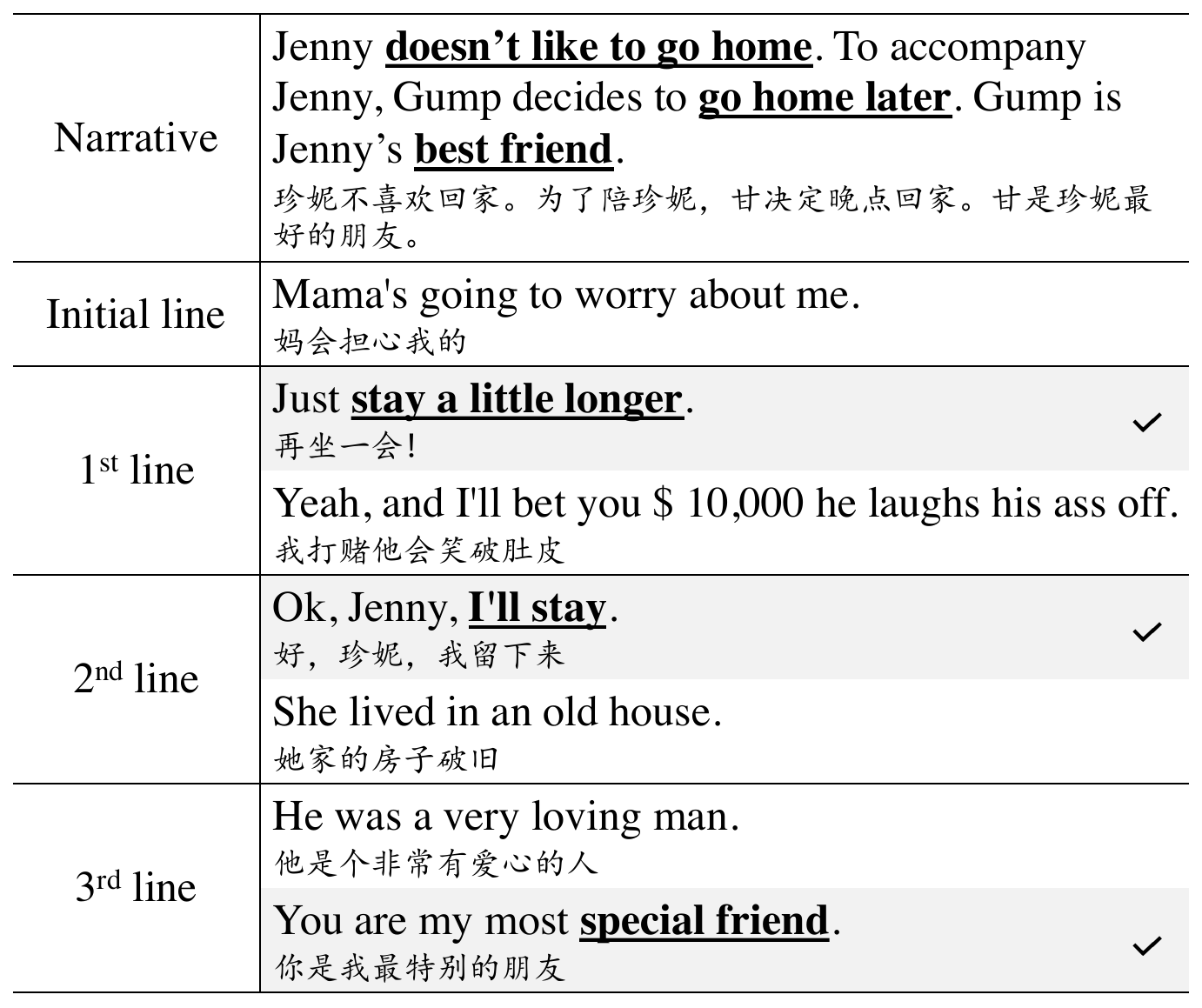}
    \caption{An example of part of a script with a narrative extracted from our \textit{GraphMovie} dataset. The checked lines are from a ground-truth session, while the unchecked responses are other candidates that are relevant but not coherent with the narrative.}
    \label{fig:example}
\end{figure}

In this work, we investigate the utilization of narratives in a special case of text generation -- \textbf{movie script generation}. This special form of conversation generation is chosen due to the unavailability of the data for a more general form of application. Yet it does require the same care to leverage narratives in general conversation, and hence can provide useful insight to a more general form of narrative-guided conversation.
The dataset we use to support our study is collected from GraphMovie\footnote{\url{http://www.graphmovies.com/home/2/index.php}. Unfortunately, we find this website was closed recently.}, where an end-user retells the story of a movie by uploading descriptive paragraphs in his/her own words. More details about the dataset will be presented in Section~\ref{sec:data_collection}. An example is shown in Figure~\ref{fig:example}, where the narrative is uploaded to retell several lines of a script in a movie. Our task is to generate/select the following lines by leveraging the narrative.

Our problem is closely related to dialogue generation that takes into account the context~\cite{DBLP:conf/acl/WuWXZL17, DBLP:conf/coling/ZhangLZZL18, DBLP:conf/acl/WuLCZDYZL18}. 
However, a narrative plays a different and more specific role than a general context. In particular, a narrative may cover the whole story (a part of a script), thus a good conversation should also cover all the aspects mentioned in a narrative, which is not required with a general context.
In this paper, we propose a new model called \textbf{ScriptWriter} to address the problem of script generation/selection with the help of a narrative. ScriptWriter keeps track of what in the narrative has been said and what is remaining to select the next line by an updating mechanism. The matching between updated narrative, context, and response are then computed respectively and finally aggregated as a matching score. As it is difficult to evaluate the quality of script generation, we frame our work in a more restricted case - selecting the right response among a set of candidates. This form of more limited conversation generation - retrieval-based conversation - has been widely used in the previous studies~\cite{DBLP:conf/acl/WuWXZL17,DBLP:conf/acl/WuLCZDYZL18}, and it provides an easier way to evaluate the impact of narratives.

We conduct experiments on a dataset we collected and made publicly available (see Section 5). The experiments will show that using a narrative to guide the generation/selection of script is a much more appropriate approach than using it as part of the general context. 

Our work has three main contributions:

(1) To our best knowledge, this is the first investigation on movie script generation with a  narrative. This task could be further extended to a more general text generation scenario when suitable data are available.

(2) We construct the first large-scale data collection \textit{GraphMovie} to support research on narrative-guided movie script generation, which is made publicly accessible.

(3) We propose a new model in which a narrative plays a specific role in guiding script generation. This will be shown to be more appropriate than a general context-based approach.

\section{Related Work}
\label{sec:related}
\subsection{Narrative Understanding}
It has been more than thirty years since researchers proposed ``narrative comprehension'' as an important ability of artificial intelligence~\cite{rapaport1989cognitive}. The ultimate goal is the development of a computational theory to model how humans understand narrative texts.
Early explorations used symbolic methods to represent the narrative~\cite{turner1994minstrel,bringsjord1999artificial} or rule-based approaches to generate the narrative~\cite{DBLP:journals/jair/RiedlY10}.
Recently, deep neural networks have been used to tackle the problem~\cite{ws-2019-narrative}, and related problems such as generating coherent and cohesive text~\cite{cho-etal-2019-towards} and identifying relations in generated stories~\cite{roemmele-2019-identifying} have also been addressed. However, these studies only focused on how to understand a narrative itself (e.g., how to extract information from a narrative). They did not investigate how to utilize the narrative in an application task such as dialogue generation.

\subsection{Dialogue Systems}
Existing methods of open-domain dialogue can be categorized into two groups: retrieval-based and generation-based.
Recent work on response generation is mainly based on sequence-to-sequence structure with attention mechanism~\cite{DBLP:ShangLL15,DBLP:VinyalsL15}, with multiple extensions~\cite{DBLP:LiGBGD16,DBLP:conf/aaai/XingWWLHZM17,DBLP:conf/ijcai/ZhouYHZXZ18,DBLP:journals/corr/abs-2002-07397,DBLP:journals/ir/ZhuDNW20}.
Retrieval-based methods try to find the most reasonable response from a large repository of conversational data, instead of generating a new one~\cite{DBLP:conf/acl/WuWXZL17,DBLP:conf/acl/WuLCZDYZL18,DBLP:conf/coling/ZhangLZZL18}. In general, the utterances in the previous turns are taken together as the context for selecting the next response.
Retrieval-based methods are widely used in real conversation products due to their more fluent and diverse responses and better efficiency. In this paper, we focus on extending retrieval-based methods by using a narrative as a plan for a  session. This is a new problem that has not been studied before.

Contrary to open-domain chatbots, task-oriented systems are designed to accomplish tasks in a specific domain~\cite{DBLP:conf/interspeech/SeneffHLPSZ98,DBLP:conf/interspeech/LevinNPBBFELPRRW00,wang2011semantic,tur2011spoken}. In these systems, a dialogue state tracking component is designed for tracking what has happened in a dialogue~\cite{DBLP:journals/csl/WilliamsY07,DBLP:conf/sigdial/HendersonTY14,xu2000task}. This inspires us to track the remaining information in the narrative that has not been expressed by previous lines of conversation. However, existing methods cannot be applied to our task directly as they are usually predefined for specific tasks, and the state tracking is often framed as a classification problem. 

\subsection{Story Generation}
Existing studies have also tried to generate a story. Early work relied on symbolic planning~\cite{DBLP:conf/ijcai/Meehan77,DBLP:journals/jvca/CavazzaCM02} and case-based reasoning~\cite{DBLP:journals/jetai/PerezS01,DBLP:journals/kbs/GervasDPH05}, while more recent work uses deep learning methods. 
Some of them focused on story ending generation~\cite{peng-etal-2018-towards, DBLP:conf/aaai/GuanWH19}, where the story context is given, and the model is asked to select a coherent and consistent story ending. This is similar to the dialogue generation problem mentioned above. Besides, attempts have been made to generate a whole story from scratch~\cite{fan-etal-2018-hierarchical, DBLP:conf/acl/FanLD19}. Compared with the former task, this latter is more challenging since the story framework and storyline should all be controlled by the model. 

Some recent studies also tried to guide the generation of dialogues~\cite{DBLP:conf/acl/WuGZWZLW19,DBLP:conf/acl/TangZXLXH19} or stories~\cite{DBLP:conf/aaai/YaoPWK0Y19} with keywords - the next response is asked to include the keywords. This is a step towards guided response generation and bears some similarities with our study. However, a narrative is more general than keywords, and it provides a description of the dialogue session rather than imposing keywords to the next response. 

\begin{table}[t]
    \centering
    \small
    \caption{Statistics of \textit{GraphMovie} corpus.}
    \begin{tabular}{lrrr}
    \toprule
         & \textbf{Training} & \textbf{Validation} & \textbf{Test} \\
    \midrule
        \# Sessions & 14,498 & 805 & 806 \\
        \# Micro-sessions & 136,524 & 37,480 & 38,320 \\
        \# Candidates & 2 & 10 & 10 \\
        Min. \#turns & 2 & 2 & 2 \\
        Max. \#turns & 34 & 27 & 17 \\
        Avg. \#turns & 4.71 & 4.66 & 4.75 \\
        Avg. \#words in Narr. & 25.04 & 24.86 & 24.18 \\
    \bottomrule%
    \end{tabular}%
    \label{tab:statistics}%
\end{table}

\section{Problem Formulation and Dataset}
\subsection{Problem Formulation}
Suppose that we have a dataset $\mathcal{D}$, in which a sample is represented as $(y,c,p,r)$, where $c=\{s_{1},\cdots,s_n\}$ represents a context formed by the preceding sentences/lines $\{s_{i}\}_{i=1}^{n}$; $p$ is a predefined narrative that governs the whole script session, and $r$ is a next line candidate (we refer to it as a response); $y\in \{0,1\}$ is a binary label, indicating whether $r$ is a proper response for the given $c$ and $p$. Intuitively, a proper response should be relevant to the context, and be coherent and aligned with the narrative. Our goal is to learn a model $g(c,p,r)$ with $\mathcal{D}$ to determine how suitable a response $r$ is to the given context $c$ and narrative $p$.

\subsection{Data Collection and Construction}\label{sec:data_collection}
Data is a critical issue in research on story/dialogue generation. Unfortunately, no dataset has been created for narrative-guided story/dialogue generation.
To fill the gap, we constructed a test collection from GraphMovie, where an editor or a user can retell the story of a movie by uploading descriptive paragraphs in his/her own words to describe screenshots selected from the movie. A movie on this website has, on average, 367 descriptions. A description paragraph often contains one to three sentences to summarize a fragment of a movie. It can be at different levels - from retelling the same conversations to a high-level description. We consider these descriptions as narratives for a sequence of dialogues, which we call a session in this paper. Each dialogue in a session is called a line of script (or simply a line).

To construct the dataset, we use the top 100 movies in IMDB\footnote{\url{https://www.imdb.com/}} as an initial list. For each movie, we collect its description paragraphs from GraphMovie. Then we hire annotators to watch the movie and annotate the start time and end time of the dialogues corresponding to each description paragraph through an annotation tool specifically developed for this purpose. 
According to the start and end time, the sequence of lines is extracted from the subtitle file and aligned with a corresponding description paragraph.

As viewers of a movie can upload descriptions freely, not all description paragraphs correspond to a narrative and are suitable for our task. For example, some uploaded paragraphs express one's subjective opinions about the movie, the actors, or simply copy the script. Therefore, we manually review the data and remove such non-narrative data.
We also remove sessions that have less than two lines. Finally, we obtain 16,109 script sessions, each of which contains a description paragraph (narrative) and corresponding lines of the script. As shown in Table~\ref{tab:statistics}, on average, a narrative has about 25 words, and a session has 4.7 lines. The maximum number of lines in a session is 34. 

Our task is to select one response from a set of candidates at any point during the session. By moving the prediction point through the session, we obtain a set of micro-sessions, each of which has a sequence of previous lines as context at that point of time, the same narrative as the session, and the next line to predict. The candidates to be selected contain one ground-truth line - the one that is genuinely the next line, together with one (in the training set) or nine (in the validation/test set) other candidates retrieved with the previous lines by Solr\footnote{\url{https://lucene.apache.org/solr/}}. The above preparation of the dataset follows the practice in the literature~\cite{DBLP:conf/acl/WuWXZL17} for retrieval-based dialogue.

\section{Proposed Method: ScriptWriter}
\subsection{Overview}
A good response is required to be coherent with the previous lines, i.e., context, and be consistent with the given narrative. For example, ``Just stay a little longer'' can respond ``Mama's going to worry about me'' and it has no conflict with the narrative in Figure~\ref{fig:example}. Furthermore, as our target is to generate all lines in the session successively, it is also required that the following lines should convey the information that the former lines have not conveyed. Otherwise, only a part of the narrative is covered, and we will miss some other aspects specified in the narrative. 

We propose an attention-based model called ScriptWriter to solve the problem. ScriptWriter follows a representation-matching-aggregation framework. First, the narrative, the context, and the response candidate are represented in multiple granularities by multi-level attentive blocks. Second, we propose an updating mechanism to keep track of what in a narrative has been expressed and explicitly lower their weights in the updated narrative so that more emphasis can be put on the remaining parts. Third, matching features are extracted between different elements: between context and response to capture whether it is a proper reply; between narrative and response to capture whether it is consistent with the narrative; and between context and narrative to implicitly track what in the narrative has been expressed in the previous lines. Finally, the above matching features are concatenated together and a final matching score is produced by convolutional neural networks (CNNs) and a multi-layer perceptron (MLP).

\subsection{Representation}
To better handle the gap in words between two word sequences, we propose to use an attentive block, which is similar to that used in Transformer~\cite{DBLP:conf/nips/VaswaniSPUJGKP17}. 
The input of an attentive block consists of three sequences, namely query ($\bQ$), key ($\bK$), and value ($\bV$). The output is a new representation of the query and is denoted as AttentiveBlock($\bQ,\bK,\bV$) in the remaining parts. This structure is used to represent a response, lines in the context, and a narrative. 

More specifically, given a narrative $p=(w_1^p, \cdots, w_{n_p}^p)$, a line $s_i=(w_1^{s_i}, \cdots, w_{n_{s_i}}^{{s_i}})$ and a response candidate $r=(w_1^r, \cdots, w_{n_r}^r)$, ScriptWriter first uses a pre-trained embedding table to map each word $w$ to a $d_e$-dimension embedding $\be$, i.e., $w \Rightarrow \be$.
Thus the narrative $p$, the line $s_i$ and the response candidate $r$ are represented by matrices $\bP^0=(\be_1^p, \cdots, \be_{n_p}^p)$, $\bS_i^0=(\be_1^{s_i}, \cdots, \be_{n_{s_i}}^{{s_i}})$ and $\bR^0=(\be_1^r, \cdots, \be_{n_r}^r)$. 

Then ScriptWriter takes $\bP^0$, $\{\bS_i^0\}_{i=1}^n$ and $\bR^0$ as inputs and uses stacked attentive blocks to construct multi-level self-attention representations. The output of the $(l-1)^{th}$ level of attentive block is input into the $l^{th}$ level. The representations of $p$, $s_i$, and $r$ at the $l^{th}$ level are defined as follows:
\begin{align}
    \bP^{l} &= \text{AttentiveBlock}(\bP^{l-1}, \bP^{l-1}, \bP^{l-1}),\label{p_representation} \\
    \bS_i^{l} &= \text{AttentiveBlock}(\bS_i^{l-1}, \bS_i^{l-1}, \bS_i^{l-1}), \\
    \bR^{l} &= \text{AttentiveBlock}(\bR^{l-1}, \bR^{l-1}, \bR^{l-1}), 
\end{align}
where $l$ ranges from $1$ to $L$.

Inspired by a previous study~\cite{DBLP:conf/acl/WuLCZDYZL18}, we apply another group of attentive blocks, which is referred to as cross-attention, to capture semantic dependency between $p$, $s_i$ and $r$.
Considering $p$ and $s_i$ at first, their cross-attention representations are defined by:
\begin{align}
    \overline{\bP}_{s_i}^l &= \text{AttentiveBlock}(\bP^{l-1}, \bS_i^{l-1}, \bS_i^{l-1}),\\  
    \overline{\bS}_{i,p}^l &= \text{AttentiveBlock}(\bS_i^{l-1}, \bP^{l-1}, \bP^{l-1}).
\end{align}
Here, the words in the narrative can attend to all words in the line, and vice verse. In this way, some inter-dependent segment pairs, such as ``stay'' in the line and ``go home later'' in the narrative, become close to each other in the representations. Similarly, we compute cross-attention representations between $p$ and $r$ and between $r$ and $s_i$ at different levels, which are denoted as $\overline{\bP}_{r}^l$, $\overline{\bR}_{p}^l$, $\overline{\bS}_{i,r}^l$ and $\overline{\bR}_{s_i}^l$. These representations further provide matching information across different elements in the next step.

\begin{figure}[!t]
    \centering
    \includegraphics[width=\linewidth]{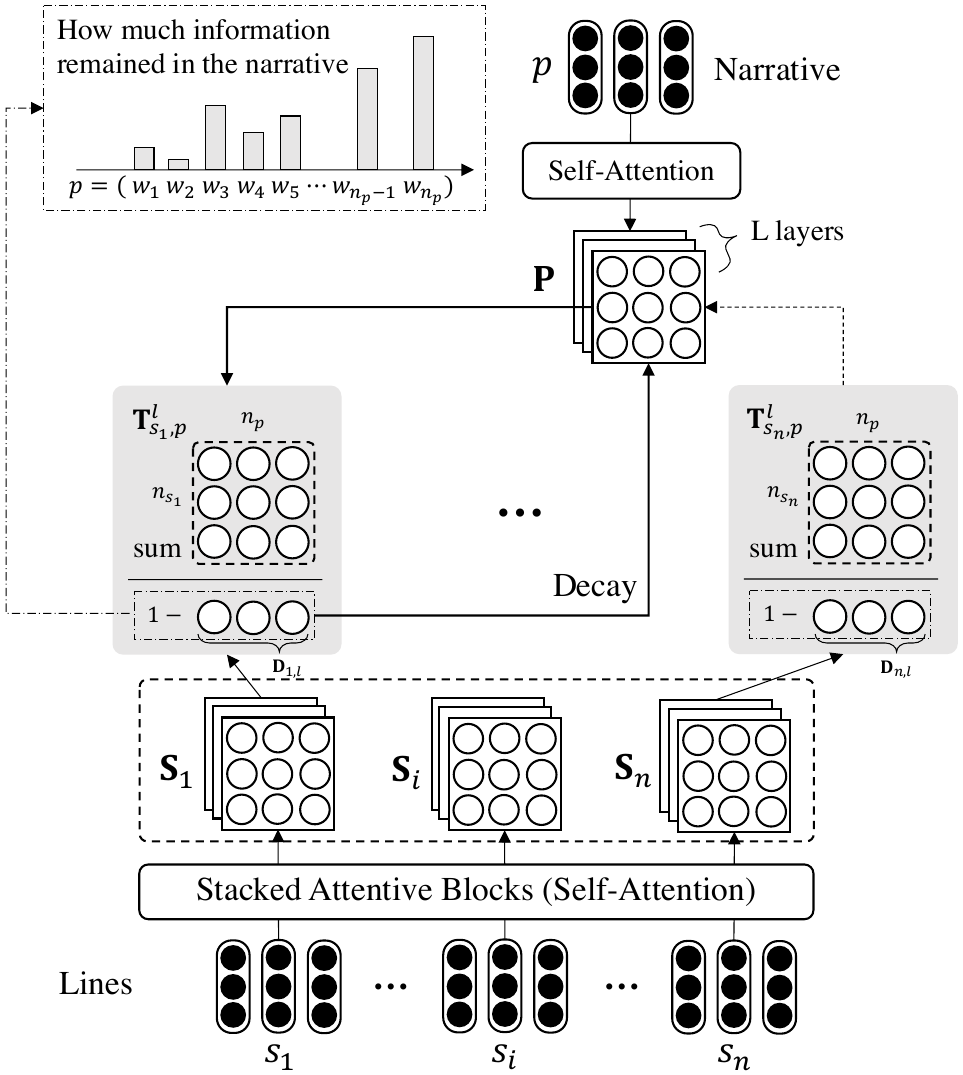}
    \caption{Updating mechanism in ScriptWriter. The representation of the narrative is updated by lines in the context one by one. The information that has been expressed is decayed. Thus the updated narrative focuses more on the remaining information.}
    \label{fig:update}
\end{figure}

\subsection{Updating Mechanism}
We design an updating mechanism to keep track of the coverage of the narrative by the lines so that the selection of the response will focus on the uncovered parts. The mechanism is illustrated in Figure~\ref{fig:update}. We update a narrative gradually by all lines in the context one by one. For the $i^{th}$ line $s_i$, we conduct a matching between $\bS_i$ and $\bP$ by their cosine similarity at all levels ($l$) of attentive blocks:
\begin{align}
    \bT_{s_i,p}^l[j][k] = \cos({\bS_i^l}[j], \bP^l[k]),
\end{align}
where $j$ and $k$ stand for the $j^{th}$ word in $s_i$ and $k^{th}$ word in $p$ respectively. To summarize how much information in $p$ has been expressed by $s_i$, we compute a vector $\bD_i$ by conducting summations along vertical axis on each level in the matching map $\bT_{s_i,p}$. The summation on the $l^{th}$ level is:
\begin{align}
    \bD_i^l &= [d_{i,1}^l,d_{i,2}^l,\cdots,d_{i,n_p}^l], \\
    d_{i,k}^l &= \gamma \sum_{j=1}^{n_{s_i}}\bT_{s_i,p}^l[j][k],
\end{align}
where $n_p$, $n_{s_i}$ denotes the number of words in $p$ and $s_i$; $\gamma \in [0,1]$ is a parameter to learn and works as a gate to control the decaying degree of the mentioned information. 
Finally, we update the narrative's representation as follows for the $i^{th}$ line $s_i$ in the context:
\begin{gather}
    \bP_{i+1}^l = (1 - \bD_i^l) \bP_i^l.
\end{gather}
The initial representation $\bP_0^l$ is equal to $\bP^l$ defined in Equation (\ref{p_representation}). 
If there are $n$ lines in the context, this update is executed $n$ times, and $(1-\bD^l)$ will produce a continuous decaying effect.

\begin{figure}[!t]
    \centering
    \includegraphics[width=\linewidth]{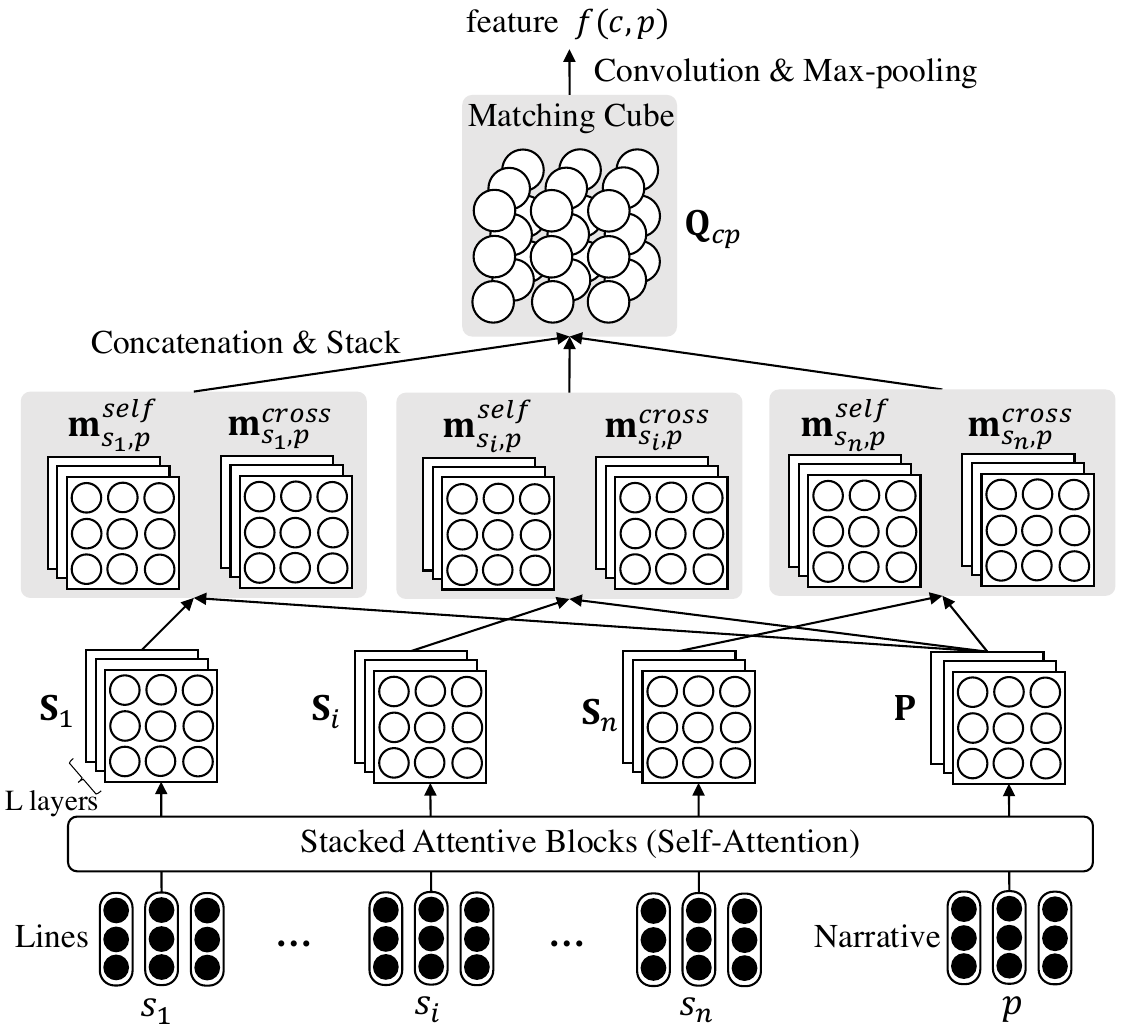}
    \caption{The context-narrative matching. All lines and the narrative are represented by attentive blocks and the matching between them results in a matching cube $\bQ_{cp}$. Matching features are aggregated and distilled by a CNN.}
    \label{fig:URmatch}
\end{figure}

\subsection{Matching}
The matching between the narrative $p$ and the line $s_i$ is conducted based on both their self-attention and cross-attention representations, as shown in Figure~\ref{fig:URmatch}. 

First, ScriptWriter computes the dot product on these two representations separately as follows:
\begin{align}
\bm_{s_i,p,l}^{self}[j,k] &= \bS_i^l[j]^T \cdot \bP^l[k],\\
\bm_{s_{i},p,l}^{cross}[j,k] &= \overline{\bS}_{i,p}^l[j]^T \cdot \overline{\bP}^l_{s_i}[k],
\end{align}
where $l$ ranges from 0 to L.
Each element is the dot product of the $j^{th}$ word representation in $\bS_i^l$ or $\overline{\bS}_{i,p}^l$ and the $k^{th}$ word representation in $\bP^l$ or $\overline{\bP}_{s_i}^l$.
Then the matching maps in different layers are concatenated together as follows:
\begin{align*}
\bm_{s_i,p}^{self}[j,k] &= \left[\bm_{s_i,p,0}^{self}\left[j,k\right] ; \cdots ; \bm_{s_i,p,L}^{self}\left[j,k\right]\right], \\
\bm_{s_i,p}^{cross}[j,k] &= \left[\bm_{s_i,p,0}^{cross}\left[j,k\right] ; \cdots ; \bm_{s_i,p,L}^{cross}\left[j,k\right]\right],
\end{align*}
where $[;]$ is concatenation operation.
Finally, the matching features computed by the self-attention representation and the cross-attention representation are fused as follows:
\begin{align*}
\bM_{s_i,p}\left[j,k\right] = \left[\bm_{s_i,p}^{self}\left[j,k\right] ; \bm_{s_i,p}^{cross}\left[j,k\right]\right].
\end{align*}
The matching matrices $\bM_{p,r}$ and $\bM_{s_i,r}$ for narrative-response and context-response are constructed in a similar way. For the sake of brevity, we omit the formulas. After concatenation, each cell in $\bM_{s_i,p}$, $\bM_{p,r}$ or $\bM_{s_i,r}$ has $2(L+1)$ channels and contains matching information at different levels.

The matching between narrative, context, and response serves for different purposes. Context-response matching ($\bM_{s_i,r}$) serves to select a  response suitable for the context. Context-narrative matching ($\bM_{s_i,p}$) helps the model ``remember'' how much information has been expressed and implicitly influences the selection of the next responses. 
Narrative-response matching ($\bM_{p,r}$) helps the model to select a more consistent response with the narrative. As the narrative keeps being updated along with the lines in context, ScriptWriter tends to dynamically choose the response that matches what remains unexpressed in the narrative.

\subsection{Aggregation}
To further use the information across two consecutive lines, ScriptWriter piles up all the context-narrative matching matrices and all the context-response matching matrices to construct two cubes $\bQ_{cp} = \{\bM_{s_i,p}[j,k ]\}_{i=1}^{n}$ and $\bQ_{cr} = \{\bM_{s_i,r}[j,k]\}_{i=1}^{n}$, where $n$ is the number of lines in the session.
Then ScriptWriter employs 3D convolutions to distill important matching features from the whole cube. We denote these two feature vectors as $f(c,p)$ and $f(c,r)$.
For narrative-response matching, ScriptWriter conducts 2D convolutions on $\bM_{p,r}$ to distill matching features between the narrative and the response, denoted as $f(p,r)$.

The three types of matching features are concatenated together, and the matching score $g(c,p,r)$ for ranking response candidates is computed by an MLP with a sigmoid activation function, which is defined as: 
\begin{align}
    f(c,p,r) &= [f(c,p) ; f(c,r) ; f(p,r)], \\
    g(c,p,r) &= \text{sigmoid}(\bW^T f(c,p,r) + b),
\end{align}
where $\bW$ and $b$ are parameters.

ScriptWriter learns $g(c,p,r)$ by minimizing cross entropy with $\mathcal{D}$. The objective function is formulated as:
\begin{align}
    L(\theta) = -&\sum_{(y, c, p, r) \in \mathcal{D}}[y\log({g(c,p,r)}) \nonumber \\
    &+(1-y)\log({1-g(c,p,r)})].
\end{align}

\section{Experiments}
\subsection{Evaluation setup}
As presented in Table~\ref{tab:statistics}, we randomly split the the \textit{GraphMovie} collection into training, validation and test set. The split ratio is 18:1:1.
We split the sessions into micro-sessions: given a session with $n$ lines in the context, we will split it into $n$ micro-sessions with length varying from 1 to $n$. These micro-sessions share the same narrative. By doing this, the model is asked to learn to select one line as the response from a set of candidates at any point during the session, and the dataset, in particular for training, can be significantly enlarged. 

We conduct two kinds of evaluation as follows:

\textbf{Turn-level task} asks a model to rank a list of candidate responses based on its given context and narrative for a micro-session. The model then selects the best response for the current turn. This setting is similar to the widely studied response selection task~\cite{DBLP:conf/acl/WuWXZL17,DBLP:conf/acl/WuLCZDYZL18,DBLP:conf/coling/ZhangLZZL18}.
We follow these previous studies and employ recall at position $k$ in $n$ candidates (R$_\text{n}$@k) and mean reciprocal rank (MRR)~\cite{DBLP:conf/trec/Voorhees99} as evaluation metrics. For example, R$_\text{10}$@1 means recall at one when we rank ten candidates (one positive sample and nine negative samples).
The final results are average numbers over all micro-sessions in the test set.

\textbf{Session-level task} aims to predict all the lines in a session gradually. It starts with the first line of the session as the context and the given narrative and predicts the best next line. The predicted line is then incorporated into the context to predict the next line. This process continues until the last line of the session is selected. Finally, we calculate precision over the whole original session and report average numbers over all sessions in the test set. Precision is defined as the number of correct selection divided by the number of lines in a session. We consider two measures: 1) P$_{\text{strict}}$ which accepts a right response at the right position; 2) P$_{\text{weak}}$ which accepts a right response at any position.

\subsection{Baselines}\label{sec:baselines}
As no previous work has been done on narrative-based script generation, no proper baseline exists. Nevertheless, some existing multi-turn conversation models based on context can be adapted to work with a narrative: the context is simply extended with the narrative. Two different extension methods have been tested: the narrative is added into the context together with the previous lines; the narrative is used as a second context. In the latter case, two matching scores are obtained for context-narrative and narrative-response. They are aggregated through an MLP to produce a final score. This second approach turns out to perform better. Therefore, we only report the results with this latter method\footnote{We also tested some basic models such as RNN, LSTM, and BiLSTM~\cite{DBLP:conf/sigdial/LowePSP15} in our experiments. However, they cannot achieve comparable results to the selected baselines.}.

(1) MVLSTM~\cite{DBLP:conf/aaai/WanLGXPC16}: it concatenates all previous lines as a context and uses an LSTM to encode the context and the response candidate. 
A matching score is determined by an MLP based on a map of cosine similarity between them. A matching score for narrative-response is produced similarly.

(2) DL2R~\cite{DBLP:YanSW16}: it encodes the context by an RNN followed by a CNN. The matching score is computed similarly to MVLSTM. 

(3) SMN~\cite{DBLP:conf/acl/WuWXZL17}: it matches each line with response sequentially to produce a matching vector with CNNs. The matching vectors are aggregated with an RNN.

(4) DAM~\cite{DBLP:conf/acl/WuLCZDYZL18}: it represents a context and a response by using self-attention and cross-attention operation on them. It uses CNNs to extract features and uses an MLP to get a score. Different from our model, this model only considers the context-response matching and does not track what in the narrative has already been expressed by the previous lines, i.e., context.

(5) DUA~\cite{DBLP:conf/coling/ZhangLZZL18}: it concatenates the last line with each previous line in the context and response, respectively. Then it performs a self-attention operation to get refined representations, based on which matching features are extracted with CNNs and RNNs. 

\subsection{Training Details}
All models are implemented in Tensorflow\footnote{\url{https://www.tensorflow.org}}. Word embeddings are pre-trained by Word2vec~\cite{DBLP:journals/corr/abs-1301-3781} on the training set with 200 dimensions. We test the stack number in \{1,2,3\} and report our results with three stacks. Due to the limited resources, we cannot conduct experiments with a larger number of stacks, which could be tested in the future. 
Two 3D convolutional layers have 32 and 16 filters, respectively. They both use [3,3,3] as kernel size, and the max-pooling size is [3,3,3]. Two 2D convolutional layers on narrative-response matching have 32 and 16 filters with [3,3] as kernel size. The max-pooling size is also [3,3]. 
All parameters are optimized with Adam optimizer~\cite{DBLP:journals/corr/KingmaB14}. The learning rate is 0.001 and decreased during training. The initial value for $\gamma$ is 0.5. The batch size is 64. We use the validation set to select the best models and report their performance on the test set. The maximum number of lines in context is set as ten, and the maximum length of a line, response, and narrative sentence is all set as 50. All sentences are zero-padded to the maximum length. We also padded zeros if the number of lines in a context is less than 10. Otherwise, we kept the latest ten lines. The dataset and the source code of our model are available on GitHub\footnote{\url{https://github.com/DaoD/ScriptWriter}}.

\begin{table}[t]
    \centering
    \small
    \caption{Evaluation results on two response selection tasks: turn-level and session-level. Our ScriptWriter model is represented as SW. $\dag$ and $\star$ denote significant improvements with SW in t-test with $p\leq0.01$ and $p\leq0.05$ respectively.}
    \begin{tabular}{p{0.15\linewidth}|p{0.078\linewidth}p{0.078\linewidth}p{0.078\linewidth}p{0.09\linewidth}|p{0.078\linewidth}p{0.078\linewidth}}
        \toprule
         & \multicolumn{4}{c|}{Turn-level} & \multicolumn{2}{c}{Session-level} \\
        \midrule
        Method & R$_{\text{2}}$@1 & R$_{\text{10}}$@1  & R$_{\text{10}}$@5 & MRR & P$_{\text{strict}}$ & P$_{\text{weak}}$ \\
        \midrule
        MVLSTM & $0.651^\dag$ & $0.217^\dag$ & $0.732^\dag$ & $0.395^\dag$ & $0.198^\dag$ & $0.224^\dag$ \\
        DL2R & $0.643^\dag$ & $0.210^\dag$ & $0.638^\dag$ & $0.314^\dag$ & $0.230^\dag$ & $0.243^\dag$ \\
        SMN & $0.641^\dag$ & $0.176^\dag$ & $0.696^\dag$ & $0.392^\dag$ & $0.197^\dag$ & $0.236^\dag$ \\
        DAM & $0.631^\dag$ & $0.240^\dag$ & $0.733^\dag$ & $0.408^\dag$ & $0.226^\dag$ & $0.236^\dag$ \\
        DUA & $0.654^\dag$ & $0.237^\dag$ & $0.736^\dag$ & $0.396^\dag$ & $0.223^\dag$ & $0.251^\dag$ \\
        \midrule
        SW & \textbf{0.730} & \textbf{0.365} & \textbf{0.814} & \textbf{0.503} & \textbf{0.373} & \textbf{0.383} \\
        \midrule
        SW$_\text{static}$ & 0.723 & 0.351 & 0.801 & $0.484^\dag$ & $0.338^\dag$ & 0.366 \\
        SW-PR & $0.654^\dag$ & $0.246^\dag$ & $0.721^\dag$ & $0.398^\dag$ & $0.223^\dag$ & $0.239^\dag$ \\
        SW-CP & $0.710^\star$ & $0.326^\dag$ & $0.793^\dag$ & $0.473^\dag$ & $0.329^\dag$ & $0.352^\dag$ \\
        SW-CR & 0.725 & $0.316^\dag$ & $0.766^\dag$ & $0.466^\dag$ & $0.335^\dag$ & 0.382 \\
        \bottomrule
    \end{tabular}
    \label{tab:result}
\end{table}

\subsection{Results and Analysis}
\subsubsection{Evaluation Results}
The experimental results are reported in Table 2. The results on both turn-level and session-level evaluations indicate that ScriptWriter dramatically outperforms all baselines, including DAM and DUA, which are two state-of-the-art models on multi-turn response selection. All improvements are statistically significant ($p$-value $\leq 0.01$). DAM performs better than other baselines, which confirms the effectiveness of the self and cross attention mechanism used in this model.
The DUA model also uses the attention mechanism. It outperforms the other baselines that do not use attention. Both observations confirm the advantage of using attention mechanisms over pure RNN.

Between the two session-level measures, we observe that our model is less affected when moving from P$_{\text{weak}}$ to P$_{\text{strict}}$. This shows that ScriptWriter can better select a response in the right position. We attribute this behavior to the utilization of narrative coverage.

\subsubsection{Model Ablation}
We conduct an ablation study to investigate the impact of different modules in ScriptWriter. First, we remove the updating mechanism by setting $\gamma=0$ (i.e., the representation of the narrative is not updated but static). This model is denoted as ScriptWriter$_\text{static}$ in Table~\ref{tab:result}. Then we remove narrative-response, context-narrative, and context-response matching, respectively. These variants are denoted as ScriptWriter-PR, ScriptWriter-CP, and ScriptWriter-CR.

Model ablation results are shown in the second part of Table~\ref{tab:result}. We have the following findings: 1) ScriptWriter performs better than ScriptWriter$_\text{static}$, demonstrating the effectiveness of updating mechanism for the narrative. The optimal value of $\gamma$ is at around 0.647 after training, which means that only about 35\% of information is kept when a line conveys it. 2) In both turn-level and session-level evaluations, the performance drops the most when we remove narrative-response matching. 
This indicates that the relevance of the response to the narrative is the most useful information in narrative-guided script generation.
3) When we remove context-narrative matching, the performance drops too, indicating that context-narrative matching may provide implicit and complementary information for controlling the alignment of response and narrative. 
4) In contrast, when we remove the context-response matching, the performance also drops, however, at a much smaller scale, especially on P$_{\text{weak}}$, than when narrative-response matching is removed. This contrast indicates that narrative is a more useful piece of information than context to determine what should be said next, thus it should be taken into account with an adequate mechanism. 

\begin{figure}[t]
    \centering
    \includegraphics[width=\linewidth]{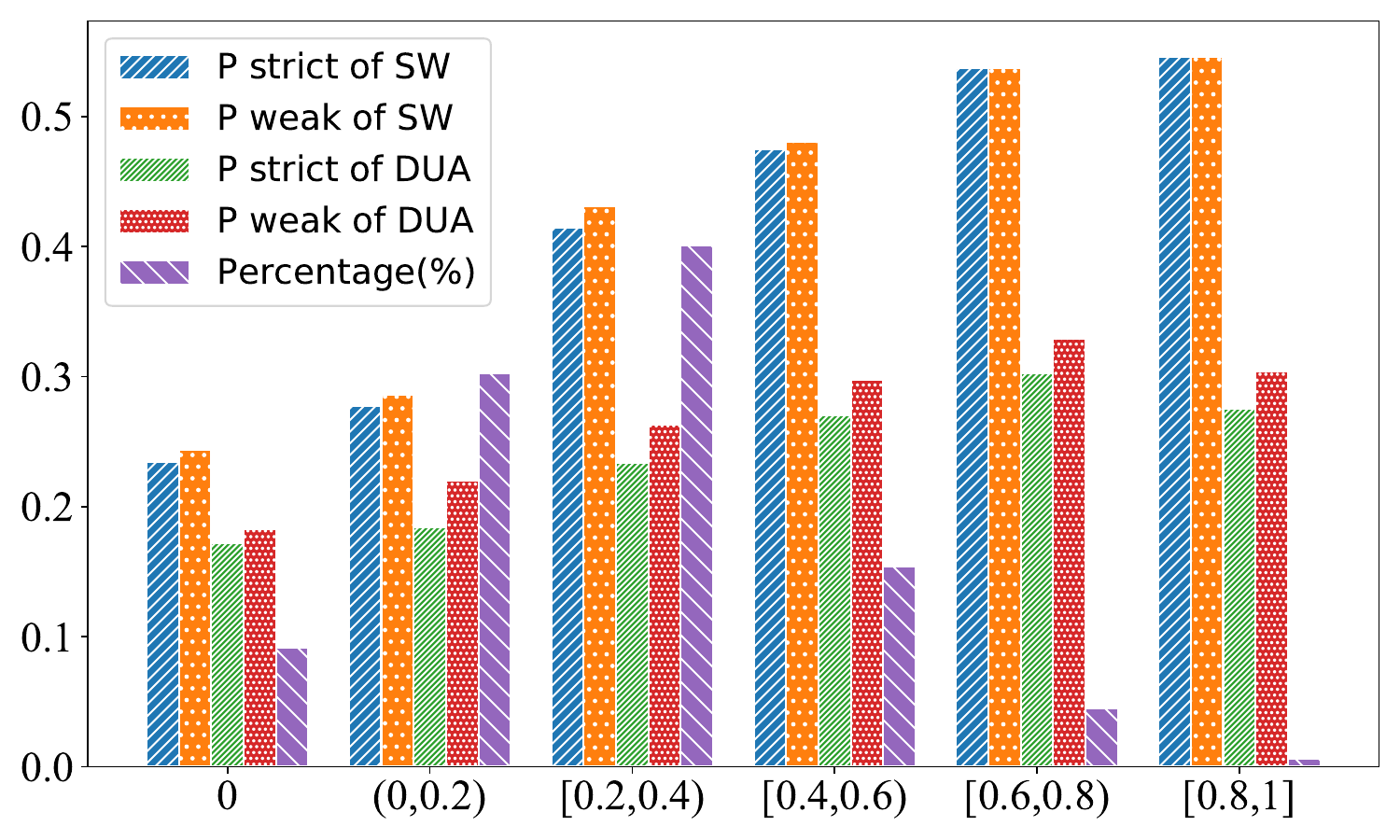}
    \caption{The performance of ScriptWriter (SW) and DUA on the test set with different types of narrative in session-level evaluation.}
    \label{fig:3}
\end{figure}

\subsubsection{Performance across Narrative Types}
\label{sec:ntype}
As we explained, narratives in our dataset are contributed by netizens, and they vary in style. Some narratives are detailed, while others are general. The question we analyze is how general vs. detailed narratives affect the performance of response selection. We use a simple method to evaluate roughly the degree of detail of a narrative: a narrative that has a high lexical overlap with the lines in the session is considered to be detailed.
Narratives are put into six buckets depending on their level of detail, as shown in Figure~\ref{fig:3}. 

We plot the performance of ScriptWriter and DUA in session-level evaluation over different types of narratives. The first type ``0'' means no word overlap between narrative and dialogue sessions. This is the most challenging case, representing extremely general narratives. It is not surprising to see that both ScriptWriter and DUA performs poorly on this type compared with other types in terms of P$_\text{strict}$. The performance tends to become better when the overlap ratio is increased. This is consistent with our intuition: when a narrative is more detailed and better aligned with the session in wording, it is easier to choose the best responses. This plot also shows that our ScriptWriter can achieve better performance than DUA on all types of narratives, which further demonstrates the effectiveness of using narrative to guide the dialogue.

We also observe that the buckets ``[0, 0.2)'' and ``[0.2, 0.4)'' contain the largest proportions of narratives. This indicates that most netizens do not use the original lines to retell a story. The problem we address in this paper is thus non-trivial. 

\section{Conclusion and Future Work}
Although story generation has been extensively studied in the literature, no existing work addressed the problem of generating movie scripts following a given storyline or narrative. In this paper, we addressed this problem in the context of generating dialogues in a movie script.
We proposed a model that uses the narrative to guide the dialogue generation/retrieval. We keep track of what in the narrative has already been expressed and what is remaining to select the next line through an updating mechanism. The final selection of the next response is based on multiple matching criteria between context, narrative and response. We constructed a new large-scale data collection for narrative-guided script generation from movie scripts.  This is the first public dataset available for testing narrative-guided  dialogue generation/selection. Experimental results on the dataset showed that our proposed approach based on narrative significantly outperforms the baselines that use a narrative as an additional context, and showed the importance of using the narrative in a proper manner. 
As a first investigation on the problem, our study has several limitations. For example, we have not considered the order in the narrative 
description, which could be helpful in generating dialogues in correct order. Other methods to track the dialogue state and the coverage of narrative can also be designed.  Further investigations are thus required to fully understand how narratives can be effectively used in dialogue generation.

\section*{Acknowledgments}
Ruihua Song and Zhicheng Dou are the corresponding authors. This work was supported by National Natural Science Foundation of China No. 61872370 and No. 61832017, and Beijing Outstanding Young Scientist Program NO. BJJWZYJH012019100020098.

\end{document}